\pdfoutput=1

\documentclass[11pt]{article}

\usepackage{EMNLP2022}

\usepackage{times}
\usepackage{latexsym}

\usepackage[T1]{fontenc}

\usepackage[utf8]{inputenc}

\usepackage{microtype}

\usepackage{inconsolata}
\usepackage{graphicx}
\usepackage{enumitem}
\usepackage{amsmath}
\usepackage{amsfonts}
\usepackage{relsize}
\usepackage[ruled,linesnumbered]{algorithm2e}
\usepackage{booktabs}
\usepackage{float}
\usepackage{natbib}
\usepackage{hyperref}

\title{ProtSi: Prototypical Siamese Network with Data Augmentation for Few-Shot Subjective Answer Evaluation}

\author{Yining Lu \and Jingxi Qiu \and Gaurav Gupta\\
AI-LAB \\
Wenzhou Kean University \\
\texttt{\{yiningl, qiujing, ggupta\}@kean.edu}}

\begin{document}
\maketitle
\begin{abstract}
Subjective answer evaluation is a time-consuming and tedious task, and the quality of the evaluation is heavily influenced by a variety of subjective personal characteristics. Instead, machine evaluation can effectively assist educators in saving time while also ensuring that evaluations are fair and realistic. However, most existing methods using regular machine learning and natural language processing techniques are generally hampered by a lack of annotated answers and poor model interpretability, making them unsuitable for real-world use. To solve these challenges, we propose ProtSi Network, a unique semi-supervised architecture that for the first time uses few-shot learning to subjective answer evaluation. To evaluate students' answers by similarity prototypes, ProtSi Network simulates the natural process of evaluator scoring answers by combining Siamese Network which consists of BERT and encoder layers with Prototypical Network.  We employed an unsupervised diverse paraphrasing model ProtAugment, in order to prevent overfitting for effective few-shot text classification. By integrating contrastive learning, the discriminative text issue can be mitigated. Experiments on the Kaggle Short Scoring Dataset demonstrate that the ProtSi Network outperforms the most recent baseline models in terms of accuracy and quadratic weighted kappa.
\end{abstract}

\section{Introduction}
The examination is a crucial method for evaluating the academic performance of students, whether objectively or subjectively. Using pre-defined correct answers, it is simple to evaluate objective multiple-choice questions. Subjective questions, on the other hand, allow students to provide descriptive responses, from which the evaluator can determine the students's level of knowledge and assign grades based on their opinion. \\
In general, subjective responses are usually lengthier and include far more substance than objective answers. Therefore, it takes a lot of effort and time for evaluators to manually assess subjective responses. In addition, the marks assigned to the same subjective questions can be varied from evaluator to evaluator, depending on their ways of evaluating, moods at the evaluating moment, and even their relationship with students \cite{bashir2021subjective}. Therefore, automated evaluation for subjective answers with a consistent process becomes necessary. Not only can the human efforts in this repetitive task be saved and spent on other more meaningful education endeavors, but the evaluation results will also be fair and plausible for students. Due to the recent Covid-19 pandemic, most universities and colleges have shifted their examinations to the online mode. Therefore, developing a subjective answer evaluation model dovetails with the actual needs of the school and has vast application value. \\
Subjective answer evaluation is a sub-field of text classification involving comparing student answers with the model answer and classifying the result into grade classes. Because questions are domain-specific, automatically evaluating answers is a challenging task because of the manually labeled data scarcity and heterogeneity of text \citep{iwata2020meta}. \\
Most of the proposed evaluation methods \citep{Patil2018SubjectiveAE, Sakhapara2019SubjectiveAG, bhonsle2019adaptive, johri2021assess, abhishek2021IRJET, bhonsle2019adaptive} rely on traditional statistical and NLP techniques. Their evaluation methods consist of four main parts: preprocessing (tokenization, stopwords removal, chunking, POS tagging, lemmatizing), vectorizing (BoW, TF-IDF, Word2Vec, LSA), computing similarity score (cosine similarity), and classifying (KNN). However, these preprocessing steps are time-consuming and cumbersome to achieve significant results, especially for long texts which are heterogeneous and noisy. Since the invention of BERT \citep{BERT} and transfer learning in NLP \citep{howard2018universal, radford2018improving}, recent deep learning models, which are robust and easier to train, outperform most approaches in NLP tasks with minimal feature engineering. Thus, there is still much room to improve existing automatic evaluation methods using advanced deep learning methods. Inspired by that, the efforts \citep{deepnlp} attempted to test the performance of CNN, LSTM, and BERT on shot answer scoring tasks, but without other model structure design and fine-tune. Additionally, the training dataset only has 1700 examples per question prompt, which is insufficient to train a deep neural network. Its highest accuracy comes from BERT and is only 71\%. Recent research \citep{bashir2021subjective} proposes a 2-step approach to evaluate descriptive questions using machine learning and natural language processing, and \citet{Sakhapara2019SubjectiveAG} applies information gained in computing the final score. Despite the effectiveness of these methods, \citet{bashir2021subjective} hired 30 annotators to prepare sufficient labeled data that requires high cost and is time-consuming in a real-world application, and \citet{Sakhapara2019SubjectiveAG} needs a large pre-graded training dataset to compute the score instead of using the model answer. \\
To handle the problem of limited labeled data, we first consider subjective answer evaluation as a meta-learning problem: the model is trained to classify students' answer scores from a consecutive set of small tasks called episodes. Each episode contains support and query two sets with a limited number of $N$ classes and a limited number of $K$ labeled data for each class. That is usually known as $N$-way $K$-shots setup in meta-learning. Recently, meta-learning has been widely applied in NLP tasks, including emotion recognition in conversation \citep{guibon2021few}, named entity recognition \citep{das2021container} or hypernym detection \citep{yu-etal-2020-hypernymy}. The score evaluating process must involve verifying students' answers with model answers (answers from teachers). This step perfectly matches the function of the Siamese network commonly used in the Computer Vision field for image matching \citep{7899663}. We adopt the Siamese network to compare students' and model answers, then output a similarity vector for further classification. There are several metric-based meta-learning algorithms: Matching network \citep{NIPS2016_90e13578} finds the weighted nearest neighbors; Prototypical Networks \citep{snell2017prototypical} computes the class representations (prototypes) and classifies query based on Euclidean distance; Relation network \citep{sung2018learning} replaces the Euclidean distance by the deep neural network. In this work, we design a Prototypical network-based model, because it is the most effective when training data is insufficient in reality \citep{pmlr-v130-al-shedivat21a} and it is the state-of-the-art for some NLP tasks, such as intent detection \citep{dopierre-etal-2021-neural}, when equipped with a BERT text encoder. \\
One challenge from meta-learning is that the models can easily overfit the biased distribution introduced by a few training examples \citep{ippolito-etal-2019-comparison}. Inspired by the recent work PROTAUGMENT \citep{dopierre2021protaugment}, we utilize a pre-trained paraphrasing model that generates various paraphrases for the original randomly sampled text as data augmentation. Through data augmentation, we can incorporate more knowledge into the model by computing an unsupervised loss, thus making the model more robust. Another common challenge in subjective answer evaluation is the discriminative text problem which indicates two answers might have similar semantic representations but belong to different classes. Through data augmentation, we can tackle this issue by contrastive learning. The followings are the main contributions of this paper:
\begin{enumerate}
    \item To the best of our knowledge, our work is the first of its kind to attempt to apply meta-learning to the evaluation of subjective answers.  
    \item We propose a novel semi-supervised approach, ProtSi Network, for evaluating students' answers from model answers.
    \item We utilize the data augmentation to deal with overfit and heterogeneous text problems.  
    \item Through extensive experimental analysis, we demonstrate that the proposed approaches outperform the state-of-the-art on questions from the Kaggle dataset.
\end{enumerate}

\section{Problem Formulation}
The meta-learning of few-shot text classification aims to learn from a set of small tasks or episodes and then transfer the learned knowledge to the testing tasks which are unseen during training. \\
\begin{figure*}[t]
     \centering
     \includegraphics[width=6.5in]{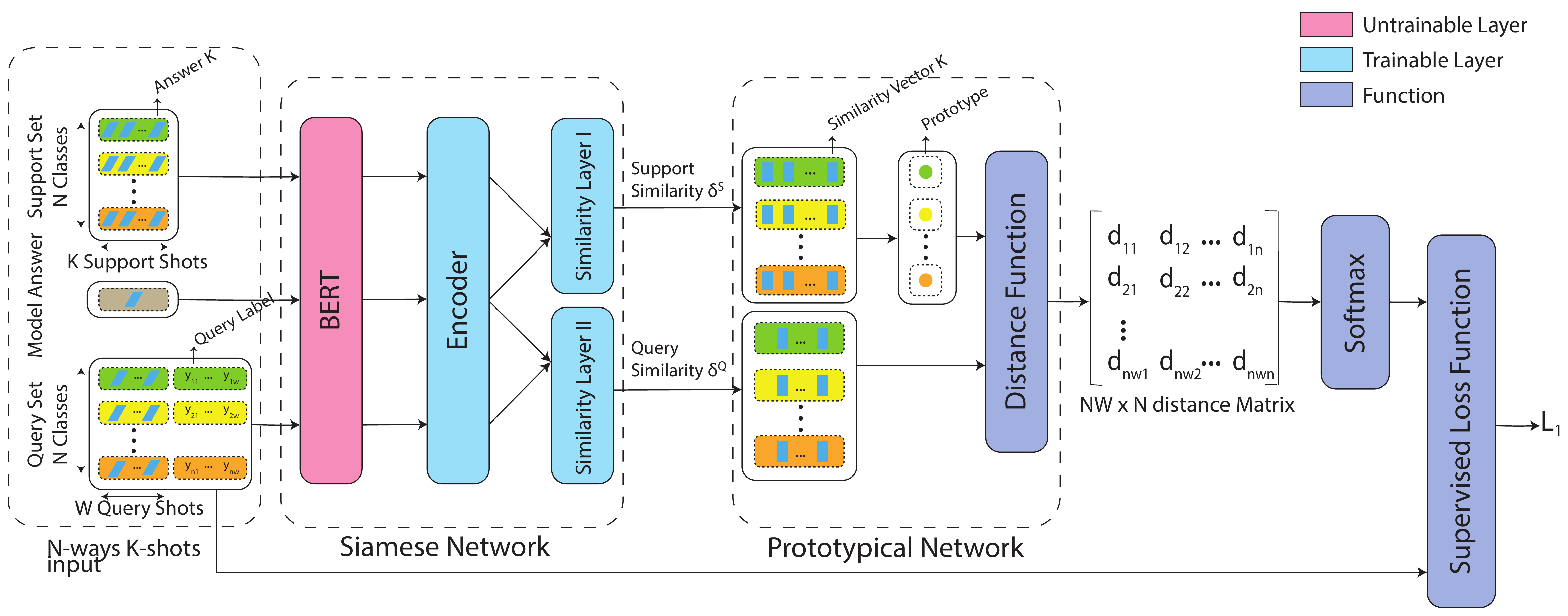}
     \caption{Framework of Supervised ProtSi Network}
    \label{supervised_protsi}
\end{figure*}
Let the student answer dataset \textit{D} be split into disjoint $\textit{D}_{\textit{train}}$, 
$\textit{D}_{\textit{val}}$,
$\textit{D}_{\textit{test}}$. At each episode, a task consists of support set $S$ and query set $Q$ is drawn from either $\textit{D}_{\textit{train}}$, $\textit{D}_{\textit{val}}$, and
$\textit{D}_{\textit{test}}$ for training, validation, or testing. In one episode, $N$ classes are randomly sampled from score class set (called rubric range). For each of the $N$ classes, $K$ students' answers are randomly sampled to compose the support set; and $W$ different student's answers are sampled to compose the query set. Let a pair $(x^{S}_{i}, y^{S}_{i})$ denote the $i^{th}$ item of total $N\times K$ items in support set. $(x^{Q}_{i}, y^{Q}_{i})$ with similar meaning denotes query set. \\
The problem becomes training a meta-learner through consecutive small tasks that attempt to classify the students' answers in the query set $Q$ based on the given model answer $m$ and a few scored answers in the support set $S$.
\section{Methodology}
In this section, we present our semi-supervised model ProtSi Network. In order to assess student answers based on similarity prototypes, ProtSi Network combines the Siamese network, which is constituted of a BERT and encoder layer, with the Prototypical network. It also incorporates unsupervised paraphrasing loss and unsupervised contrastive loss to address the overfitting problem and discriminative text problem. 
\subsection{Supervised ProtSi Network}
\subsubsection{Siamese network}
For a few-shot subjective answer evaluation, before assigning marks to students' answers, a comparison between students' answers and a model answer $M$ is needed to map the textual and contextual difference into a similarity vector so that the further metrics for classification can be computed. Unlike regular Siamese network consisting of twin networks that accept two distinct inputs \citep{koch2015siamese}, our Siamese network accepts three inputs $x^{S}_{nk}, x^{Q}_{nw}, M$ and performs dual comparisons for $M$ with $x^{S}_{nk}$ and $x^{Q}_{nw}$ separately, where $x^{S}_{nk}$ and $x^{Q}_{nw}$ are answers from support set and query set. \\
Specifically, in Siamese network, a text embedding layer is needed to map the textual data onto a high dimensional vector space where the similarity function between texts can be applied. Recently, more and more NLP tasks utilize pre-trained language models, such as BERT, to obtain text representations and  achieve promising results. Following the work \citep{chen2022contrastnet, dopierre2021protaugment, hui2020few, mehri2020natural}, we also use the pre-trained $BERT_{base}$ \citep{bert-base} as the embedding layer. The BERT model encodes the word information more effectively and flexibly because the context is considered. Thus, the numeric representations of a word can differ among sentences. For later use, we denote the $BERT_{base}$ embedding as function $e(x)$. \\
After obtaining word representation, a context encoder layer is needed to extract the whole sentence information and outputs a sentence feature vector. A recurrent neural network (RNN) is widely used in the NLP field because it can memorize the information from prior inputs and influence the current inputs and outputs. We express the embedding and encoding operations, including both the BERT and encoder layers, as the following equation:
\begin{equation}
x^{\prime} = f(e(x)|\theta)
\end{equation}
where $\theta$ denotes the trainable parameters in the encoding function $f(\cdot):\mathbb{R}^{E}\to\mathbb{R}^{M}$, which convert the E-dimensional input $x$ to the M-dimensional encoded output $x^{\prime}$. To be noticed, support and query set share the same BERT and Encoder layers. The encoded support and query set are $x^{\prime S}_{\:\:i}$ and $x^{\prime Q}_{\:\:i}$ and encoded model answer is $m^{\prime}$. Then we use two different similarity layers $s_1(\cdot,\cdot|\theta_1), s_2(\cdot,\cdot|\theta_2): \mathbf{R}^M\to \mathbf{R}^M$ to draw the similarity representations between model answer with support set and query set separately. The similarity representations of answers in support set and query set can be computed as:
\begin{align*}
    \delta^{S}_i &= s_1(x^{\prime S}_{\:\:i}, m^{\prime}|\theta_1) \\
    \delta^{Q}_i &= s_2(x^{\prime Q}_{\:\:i}, m^{\prime}|\theta_2)
\end{align*}
The cardinality of set $\{\delta^{S}_i\}$ is $NK$ and of set $\{\delta^{Q}_i\}$ is $NW$.
\subsubsection{Prototypical network}
The Prototypical network \citep{snell2017prototypical} is a few-shot classification model that computes prototypes for each class in the support set and assigns samples in the query set to classes based on their distances with prototypes. The prototypes are computed by averaging all the instance representations belonging to the same class in the support set. For our ProtSi Network with N-ways K-shots input, the prototypes are computed as:
\begin{equation}
\mathbf{c}_n = \frac{1}{K}\sum_{k=1}^{K}\delta^{S}_{nk}
\label{supervised_prototypes}
\end{equation}
where $\delta_{nk}^{S}\in\mathbb{R}^{M}$ is the similarity vector of answers $k$ in the support set class $n$ with model answer and similarity prototype $\mathbf{c}_{n}\in\mathbb{R}^{M}$ is the M-dimensional representation for class $n$. Given a distance function $d(\cdot,\cdot):\mathbb{R}^{M}\times \mathbb{R}^{M}\to\mathbb{R}^{1}$, we can compute the distance $d(\mathbf{c}_n,\delta^{Q}_{i})$ between prototypes $\mathbf{c}_n$ and the $i^{th}$ similarity vectors in the query set. Then the probability over classes for a query answer $x^{Q}_{i}$ based on a softmax over distances to the 
prototypes can be computed as the following equation:
\begin{equation}
\label{supervised_probability}
p(y=n|x^{Q}_{i}) = \frac{exp(-d(\mathbf{c}_n, \delta^{Q}_{i}))}{\sum_{j=1}^{N}exp(-d(\mathbf{c}_j, \delta^{Q}_{i}))}   
\end{equation}
Given the computed probability distribution $p_{i}$ over score classes for the $i^{th}$ answer in the query set and the one-hot truth label $y^{Q}_{i}$, the supervised cross entropy loss can be computed as:
\begin{equation}
\label{loss1}
    \mathcal{L}_1 = -\frac{1}{NW}\sum_{i=1}^{NW} y^{Q}_{i}\cdot log(p_{i})
\end{equation}
where the inner product is used.
The supervised ProtSi Network is shown in Figure \ref{supervised_protsi}.
\subsection{Unsupervised Regularization}
\begin{figure*}[t]
     \centering
     \includegraphics[width=6.5in]{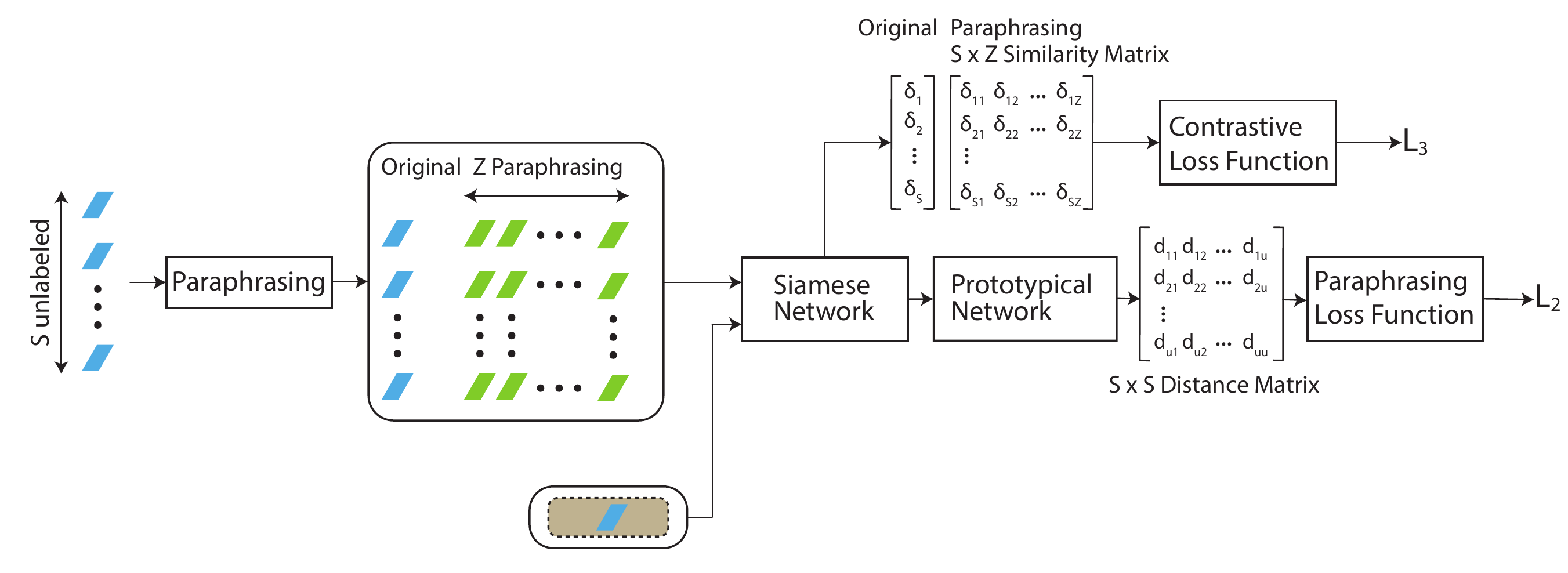}
     \caption{Framework of Unsupervised ProtSi Network}
    \label{paraphrasing_loss}
\end{figure*}

To alleviate the overfitting problem caused by a few training examples in one episode and the discriminative text problem, we propose to train the supervised ProtSi Network under the regularization of unsupervised paraphrasing loss and unsupervised contrastive loss. The overall unsupervised part of ProtSi Network is shown in Figure \ref{paraphrasing_loss}.

\subsubsection{Data Augmentation}
Along with the labled answers randomly sampled at each episode, we also randomly choose $S$ unlabeled answers from the whole dataset, including $D_{train}, D_{val}, D_{test}$. Then in the data augmentation step, we obtain $Z$ paraphrases for each unlabeled answer. We use $x_{s}$ to denote the $s^{th}$ sampled unlabeled answer in the unlabeled sample set $U = \{x_1, x_2, \cdots, x_S\}$ and $\tilde{x}^{z}_{s}$ to denote the $z^{th}$ paraphrase of $x_{s}$. The recent work \citep{dopierre2021protaugment} proposes a diverse paraphrasing model PROTAUGMENT for data augmentation and it has shown to be effecitve in few-shot text classification. We apply PROTAUGMENT to do data augmentation in our unsupervised learning.
\subsubsection{Paraphrasing Loss}
After generating paraphrases for each unlabeled answer, we put the original answer, paraphrases, and model answer through the same ProtSi Network as shown in Figure \ref{paraphrasing_loss} to create unlabeled similarity prototypes. Specifically, let $\delta_{s}$ denote the similarity vector of unlabeled answer $x_s$ with model answer and $\tilde{\delta}^{z}_{s}$ denote the similarity vector of the $z^{th}$ paraphrase of $x_{s}$. In Prototypical network, we derive unlabeled similarity prototypes of $x_{s}$ by averaging all similarity vectors of its paraphrases:
\begin{equation}
\label{unsupervised_prototypes}
    \mathbf{c}_{x_s} = \frac{1}{Z}\sum_{z=1}^{Z}\tilde{\delta}^{z}_{s}
\end{equation}
After obtaining the unlabeled similarity prototypes, we can compute the distance between $\delta_{s}$ and $\mathbf{c}_{x_s}$ using the function $d(\cdot,\cdot)$. Given the computed $S\times S$ distance matrix, we model the probability of each unlabeled similarity vector being assigned to each unlabeled similarity prototype as the Equation \ref{supervised_probability} but unsupervised:
\begin{equation}
\label{unsupervised_probability}
    p(s=t|x_s) = \frac{exp(-d(\mathbf{c}_{x_t}, \delta_{s}))}{\sum_{x_s\in U}exp(-d(\mathbf{c}_{x_s}, \delta_{s}))}
\end{equation}
Given the probability distribution $p_{x_s}$ of the unlabeled sample be assigned to unlabeled prototypes, we can compute an unsupervised entropy loss $\mathcal{L}_2$ using the following equation as the entropy regularization \citep{mnih2016asynchronous}:
\begin{equation}
\label{loss2}
    \mathcal{L}_2 = -\frac{1}{S}\sum_{x_s\in U}p_{x_s}\cdot log(p_{x_s} + \beta)
\end{equation}
where $\beta$ is a hyperparameter to prevent $log(p_{x_s})$ being negative infinity when $p_{x_s}$ is close to zero. $\mathcal{L}_2$ encourages the answers' similarity representation closer to its paraphrases' similarity prototypes and further away from the prototypes of other unlabeled similarity representations. According to \citet{mnih2016asynchronous} and \citet{dopierre2021protaugment}, adding this unsupervised paraphrasing loss improves the model's exploration ability. It brings more knowledge and noise from unlabeled answers to the model, thus making it more robust to alleviate the overfitting problem.

\subsubsection{Contrastive Loss}
Classifying discriminative text is a big challenge in text classification tasks. A Prototypical network is unsatisfactory in learning discriminative texts with similar semantics and representations belonging to different classes. It simply learns the representations but ignores the relationship among texts. To tackle the discriminative text problem in our ProtSi Network, we propose to use unsupervised contrastive loss that encourages learning discriminative text representations via contrastive learning motivated by its successful application in the few-shot text classification
\citep{chen2022contrastnet}. \\
After getting the similarity matrix $\mathcal{D} \in \mathbb{R}^{S\times Z}$ of paraphrases from the Siamese network, we randomly sample $L$ columns from $\mathcal{D}$ so that each original similarity vector $\delta_s$ is paired with $l$ new similarity vectors from paraphrasing. We combine all $\delta_s$ and sampled columns as a training batch $\{\delta\}$ of total $S(L+1)$ similarity vectors. $\delta_i$ denotes the $i^{th}$ similarity vector and $\delta^{\prime}_{i}$ denotes the paired similarity vector of $\delta_i$ in the batch $\{\delta\}$. The contrastive loss is:
\begin{equation}
\label{loss3}
    \mathcal{L}_3 {=} -\frac{1}{\scriptstyle S(L+1)}\mathlarger{\sum}_{\delta_i \in \{\delta\}}\footnotesize log\frac{ \sum\limits_{\delta_i=\delta^{\prime}_i}exp(\frac{\delta_i\cdot \delta^{\prime}_i}{\tau})}{ \sum\limits_{\delta_i=\delta^{\prime}_i}exp(\frac{\delta_i\cdot \delta^{\prime}_i}{\tau})+\sum\limits_{\delta_i\neq\delta^{\prime}_i}exp(\frac{\delta_i\cdot \delta^{\prime}_i}{\tau})}
\end{equation}
Each similarity representation is paired and compared to all other representations. Thus the contrastive loss $\mathcal{L}_3$ encourages different similarity representations to be distant from each other and pull closer similarity representations belonging to the same original answers, which prevents the different answers with similar representations from being assigned the same score and alleviate the discriminative text problem.

\subsubsection{Final Objective}
The supervised Prototypical network in Figure \ref{supervised_protsi} aligns the query similarity vectors to the similarity prototypes and generates an $NW\times N$ distance matrix by $d(\cdot,\cdot)$. Then the score of query answer $x^Q_i \in Q$ is determined as the class of prototypes $\mathbf{c}_n$ who has the minimum distance with $x^Q_i$:
\begin{equation}
\label{prediction}
    \hat{y}^Q_i = \arg \min_{n} d(\mathbf{c}_n, \delta^Q_i)
\end{equation}
During training, we combine supervised loss $\mathcal{L}_1$ and unsupervised loss $\mathcal{L}_2$ and $\mathcal{L}_3$ into the final loss $\mathcal{L}$ using the following equation:
\begin{equation}
\label{loss}
    \mathcal{L} = t^{\alpha} \times \mathcal{L}_1 + (1-t^{\alpha}) \times \mathcal{L}_2 + \gamma \times \mathcal{L}_3
\end{equation}
where $\alpha$ and $\gamma$ are hyperparameters and $t$ is the monotonically decreasing function with respect to the epoch. We use a loss annealing scheduler which will gradually incorporate the noise from the unsupervised loss $\mathcal{L}_2$ as training progresses. \\
To summarize, the final loss mainly uses the supervised loss so that the model learns to classify. Then incorporating more and more noise and knowledge from paraphrasing data using paraphrasing loss makes the model more robust to unseen answers. Finally, the contrastive loss is introduced to learn more distinct similarity representations of different answers to deal with discriminative text problems.

\subsection{Overall Algorithm}
The overall algorithm is shown in Algorithm \ref{algorithm}.
\begin{algorithm*}[h]
\caption{ProtSi Network}
\label{algorithm}
\KwData{Model answer $m$, Student answer dataset $D$}
\KwIn{EPOCHS, EPISODES, $N$, $K$, $W$, $S$, $Z$, $L$}
\KwOut{Predicted score $\hat{y}^Q_i$}
\tcp{Prepare Dataset}
Splite $D$ into disjoint $D_{train}$, $D_{val}$, $D_{test}$\;
\For{episode$\leftarrow0$ \KwTo EPISODES}{
Ransomly sample $N$-ways $K$-shots support set $S$ and $N$-ways $W$-shots query set $Q$ from either $D_{train}$, $D_{val}$, or $D_{test}$ to create training dataset, validation dataset and testing dataset, respectively\;
Randomly sample $S$ unlabeled answers from the entire dataset $D$\;
Obtain $Z$ paraphrases from the pre-trained paraphrasing model for each unlabeled answer\; 
}

\tcp{Training}
\For{epoch$\leftarrow 0$ \KwTo EPOCHS}{
\For{$x^S_i, x^Q_i, y^Q_i, x_s, \tilde{x}^z_s, m$ \textbf{in} training dataset, unlabel answer, paraphrases, model answer}{
$\delta^{S}_i \leftarrow s_1(f(e(x^S_i)|\theta_0), f(e(m)|\theta_0)|\theta_1)$\;
$\delta^{Q}_i \leftarrow s_2(f(e(x^Q_i)|\theta_0), f(e(m)|\theta_0)|\theta_2)$\;
$\delta_s \leftarrow s_1(f(e(x_s)|\theta_0),
f(e(m)|\theta_0)|\theta_1)$ \tcp{either $s_1$ or $s_2$}
$\tilde{\delta}^z_s \leftarrow s_1(f(e(\tilde{x}^z_s)|\theta_0), f(e(m)|\theta_0)|\theta_1)$ \tcp{either $s_1$ or $s_2$}
$\hat{y}^Q_i$ $\leftarrow$ $\delta^{S}_i$, $\delta^{Q}_i$ using equations \ref{supervised_prototypes}, \ref{prediction}\;
$\mathcal{L}_1$ $\leftarrow$ $\delta^{S}_i$, $\delta^{Q}_i$, $y^Q_i$ using equations \ref{supervised_prototypes}, \ref{supervised_probability}, \ref{loss1}\;
$\mathcal{L}_2$ $\leftarrow$ $\delta_s$, $\tilde{\delta}^z_s$ using equations \ref{unsupervised_prototypes}, \ref{unsupervised_probability}, \ref{loss2}\;
Randomly sample $L$ columns from the matrix $\mathcal{D}$ of $\tilde{\delta}^z_s$ and combine with $\delta_s$ as $\{\delta\}$\;
$\mathcal{L}_3$ $\leftarrow$ $\{\delta\}$ using equation \ref{loss3}\;
$\mathcal{L}$ $\leftarrow$ $\mathcal{L}_1$, $\mathcal{L}_2$, $\mathcal{L}_3$ using equation \ref{loss}\;
Perform backpropagation
}
}
\tcp{Testing}
Perform the same steps from line 9-18 and compute accuracy and quadratic weighted kappa
\end{algorithm*}
\section{Experiment}
\subsection{Datasets}
We used the Short Answer Scoring dataset by the Hewlett Foundation from Kaggle. The dataset consists of 10 question prompts, which are from different domains. Questions 1, 2, 10 are from the Science subject, Questions 5, 6 are from Biology subject, Questions 7, 8, 9 are from English subject, and Questions 3, 4 are from English Language Art subject. There are around 1700 labeled answers per question with an average length of 50 words. Different questions have different model answers and rubric ranges.

\subsection{Baselines}
We compare our method with the following competitive baseline methods released in the recent two years\footnote{Because the source code of \textbf{BERT}, \textbf{IG\_WN}, \textbf{ASSESS}, and \textbf{SACS} are not released, we re-implemented all these four methods on the same Kaggle dataset by closely following the methods described in \citep{deepnlp}, \citep{Sakhapara2019SubjectiveAG}, \citep{johri2021assess}, and \citep{abhishek2021IRJET} respectively.}:
\begin{description}
    \item[\textbf{BERT}] \citep{deepnlp} uses BERT base model with a dense layer for score classification using the labeled answers.
    \item[\textbf{IG\_WN}] \citep{Sakhapara2019SubjectiveAG} uses information gain with WordNet to vectorize textual answers and apply KNN on the consine similarity between labeled answer dataset and input answer.
    \item[\textbf{ASSESS}] \citep{johri2021assess} proposes an Assessment algorithm consisting of semantic learning to compute the similarity score between model answer and input answer.
    \item[\textbf{SACS}] \citep{abhishek2021IRJET} proposes a \textbf{S}ubjective \textbf{A}nswer \textbf{C}hecker \textbf{S}ystem using NLP and Machine Learning to compute the final score from model anwer and input answer.
\end{description}

\subsection{Implementation Details}
In the Encoder layer, we use three LSTM networks to extract the context information of sentences. Moreover, the encoded student's answer and encoded model answer are concatenated to form the input of similarity layer 1 or similarity layer 2. Each similarity layer has one dense layer with a Relu activation function and is trained using batch normalization. We choose Euclidean distance as the distance function in the prototypical network. \\
To demonstrate the efficiency of the proposed framework, we generated 8100 episodes of data for training and 2700 episodes for testing. For each episode or task, we set $N=4$ for those questions with a rubric range from 0 to 3 and $N=3$ for other questions ranging from 0 to 2. We set query shots $W=1$ and support shots $K=3$. Along with each episode, we randomly sample $S=5$ unlabeled answers and generate $Z=5$ paraphrases for each unlabeled answer with a 0.5 diversity penalty and 15 num beams for the Beam search algorithm. In the BERT layer, we set the maximum tokens per answer as 90 and the embedding size as 768. We set $L=1$ for computing contrastive loss and $\alpha=0.8$, $\beta=0.01$, $\gamma=0.02$ for computing the final loss.\\
We evaluate our method and all baseline methods on the same dataset using the Accuracy (Acc) and the Quadratic Weighted Kappa (QWK) metrics.

\subsection{Results}
\begin{table*}[t]
\centering
\begin{tabular}{lcccccccc|cl}\toprule
                 & \multicolumn{2}{c}{$\textbf{IG\_WN}^{\dagger}$} & \multicolumn{2}{c}{\textbf{ASSESS}} & \multicolumn{2}{c}{\textbf{SACS}} & \multicolumn{2}{c}{$\textbf{BERT}^{\dagger}$} & \multicolumn{2}{|c}{\textbf{ProtSi Network}} \\ \cmidrule(lr){2-3} \cmidrule(lr){4-5} \cmidrule(lr){6-7} \cmidrule(lr){8-9} \cmidrule(lr){10-11}
                 & Acc                 & QWK           & Acc            & QWK           & Acc          & QWK     & Acc         & QWK      & Acc  & QWK             \\ \midrule
$Q1$               & 0.5075              & 0.5438              & 0.4533              & 0.5815              & 0.3307                   & 0.4074          & 0.5938              & 0.7779          & 0.6181    & 0.9054                \\ 
$Q2$               & 0.3086              & 0.1826              & 0.3748              & 0.2907              & 0.2754                   & 0.1189          & 0.5312              & 0.5254          & 0.4819    & 0.6691                \\ 
$Q3$               & 0.4116              & 0.0724              & 0.2047              & 0.0191              & 0.2909                   & 0.0425          & 0.4062              & 0.2241          & 0.4325    & 0.4000                \\ 
$Q4$               & 0.5776              & 0.3317              & 0.2865              & 0.2262              & 0.4212                   & 0.1361          & 0.7500              & 0.6266          & 0.7193    & 0.8197                \\ 
$Q5$               & 0.8273              & 0.4652              & 0.2540              & 0.1513              & 0.4735                   & 0.2797          & 0.8750              & 0.7591          & 0.7796    & 0.9067                \\ 
$Q6$               & 0.8556              & 0.3282              & 0.3745              & 0.2192              & 0.3027                   & 0.0762          & 0.8438              & 0.7393          & 0.7028    & 0.8611                \\
$Q7$               & 0.5889              & 0.3661              & 0.3535              & 0.1676              & 0.3435                   & 0.1171          & 0.7812              & 0.6474          & 0.6535    & 0.6429                \\ 
$Q8$               & 0.4583              & 0.2565              & 0.5025              & 0.2927              & 0.4235                   & 0.2453          & 0.6562              & 0.5433          & 0.6051    & 0.6667                \\ 
$Q9$               & 0.5528              & 0.4919              & 0.4900              & 0.4101              & 0.4522                   & 0.2978          & 0.6875              & 0.7354          & 0.6627    & 0.7931                \\ 
$Q10$              & 0.4939              & 0.4071              & 0.5878              & 0.4194              & 0.5579                   & 0.3337          & 0.8438              & 0.6146          & 0.7101    & 0.7407                \\ 
\textbf{Average} & 0.5582              & 0.3446              & 0.3881             & 0.2778              & 0.3872                   & 0.2055          & 0.6968              & 0.6193          & \textbf{0.6366}    & \textbf{0.7405}                \\ \bottomrule
\end{tabular}
\caption{Performance comparison on Short Answer Scoring dataset. IG\_WN \protect\citep{Sakhapara2019SubjectiveAG}, ASSESS \protect\citep{johri2021assess}, SACS \protect\citep{abhishek2021IRJET}, and BERT \protect\citep{deepnlp} resuls are re-implemented by ourselves. $\dagger$ represents the methods that do not utilize model answer to evaluate the input answer but use labeled data with supervised model to directly predict score.}
\label{comparison}
\end{table*}
\subsubsection{Main Results}
Table \ref{comparison} summarizes the testing results of four baseline methods and our ProtSi Network on the Short Answer Scoring dataset with limited labeled data\footnote{The original works of ASSESS and SACS did not release their collected datasets. While the original works of BERT and IG\_WN have tested their methods on the same Short Answer Scoring dataset from Kaggle, and the results we re-implemented basically match their results in the paper.}. Rather than taking advantage of model answers, IG\_WN and BERT directly use labeled answers for supervised prediction, which is more like "learning for learning". Hence their Acc and QWK are generally higher than those of the other two baseline methods. However, ASSESS, SACS, and our ProtSi Network, which utilize the model answers for comparison and evaluation, are more dovetail with how real teachers grade answers. Furthermore, our ProtSi Network establishes new state-of-the-art performance on questions from different subjects, outperforming the ASSESS and SACS, which also utilize model answers by an average Acc of 24.85\%, 24.94\%, and by an average QWK of 46.27\%, 53.50\%. Compared to the 'best' baseline methods, our ProtSi Network also has a competitive performance, outperforming the IG\_WN by an average Acc of 7.84\% and an average QWK of 39.59\%, and outperforming the BERT by an average QWK of 12.12\%. \\
Even if the BERT method has the highest average accuracy in directly predicting students' scores, our ProtSi Network is better at distinguishing discriminative texts and has the most reliable prediction results, according to its highest average quadratic weighted kappa score. For example, in $Q10$, there are two answers with different scores.
\begin{itemize}
    \item[\textbf{\footnotesize Score 0}]light gray: on cold days the light gray absorbs the hotter temp. and on hot days it does not absorb as much.
    \item[\textbf{\footnotesize Score 2}]light gray: Light gray would be a little cool in tempature as the black or dark gray one would get to hot and the white one would get a little cold.
\end{itemize}
The BERT method considered these two answers having the same context and assigned them to score 1. However, our ProtiSi Network is more sensitive to the difference between the student and model answers. It realized that the second answer contained more information and precise expression, since the other colors were also taken into consideration. Therefore, our ProtSi Network marked it with a higher score and matched the correct human-made score.
\subsubsection{Ablation Study}
We conducted ablation studies on $Q10$ by removing one of the unsupervised losses.
\begin{table}[h]
    \centering
    \begin{tabular}{ccccc} \toprule
        $\mathcal{L}_1$ & $\mathcal{L}_2$ & $\mathcal{L}_3$ & Acc    & QWK    \\ \midrule
        $\checkmark$    & $\checkmark$    & $\checkmark$    & \textbf{0.7101} & \textbf{0.7407} \\ 
        $\checkmark$    & $\checkmark$    &                 & 0.6809 & 0.7018 \\
        $\checkmark$    &                 & $\checkmark$    & 0.6961 & 0.6557 \\
        $\checkmark$    &                 &                 & 0.6909 & 0.6552 \\ \bottomrule
    \end{tabular}
    \caption{Ablation study on $Q10$ by remvoing one of the unsupervised loss}
    \label{ablation}
\end{table}
As shown in Table \ref{ablation}, we found that removing $\mathcal{L}_2$ or $\mathcal{L}_3$ leads to performance droppings. Removing $\mathcal{L}_3$ results in 2.92 point and 3.89 point Acc and QWK decline, reflecting that unsupervised contrastive loss does teach the model to classify discriminative answers and make the results more consistent with the real score distribution. The unsupervised paraphrasing loss $\mathcal{L}_2$ is intended to improve the robustness of the model, and removing $\mathcal{L}_2$ leads to 1.40 point and 8.5 point Acc and QWK decrease because the model performs poorly in classifying unseen answers. Removing both $\mathcal{L}_2$ and $\mathcal{L}_3$ resulted in a more severe performance reduction than just removing single loss. 
\section{Conclusion}
In this paper, we presented the first study on subjective answer evaluation using few-shot learning. We proposed a novel semi-supervised architecture ProtSi Network following the teacher's real scoring ways. Our ProtSi Network used data augmentation with unsupervised paraphrasing loss to solve the overfitting problem and used contrastive learning to alleviate the discriminative text problem. Experimental results on the Short Answer Scoring dataset showed that ProtSi Network outperformed the current baseline methods under the limitation of labeled data and demonstrated its robustness across different subjects.

\section*{Limitations}
While the ProtSi Network seems suitable for real-world applications with a small number of labeled answers and outperforms the baseline methods, it still has limitations regarding its architecture. It uses an untrainable BERT layer to convert words to vectors without fine-tuning and might lose some performance. In ProtSi Network, we use a neural network to learn answer similarities and a simple Euclidean distance function to find the distances. While these yield better performance, they do not guarantee that the similarity of two high-dimensional vectors is correctly calculated, because the two vectors are far apart by the Euclidean distance but they may still be oriented closer together according to the Cosine similarity or other similarity measures in information retrievals.\\
Another limitation of the ProtSi Network is that it does not consider the complexity of questions. There are multiple sub-questions under one primary prompt, and each sub-question may have its scoring rubric and rubric range. That requires the model to have a better learning ability to evaluate multiple sub-questions simultaneously.

\section*{Ethics Statement}
we claim that our solution beats the performance of previous methods on a common open-source dataset. To demonstrate the improvement, We re-implement the baseline methods published in the recent three years by strictly following the papers. Our model has not achieved about 100\% testing accuracy on the public datasets, so when educational systems want to deploy our model, please test with their private dataset to ensure high testing accuracy because subjective answer evaluation requires exceptionally high accuracy to ensure the fairness of scoring.

\bibliographystyle{acl_natbib}

\appendix

\section{More Implementation Details}
The concrete model answers of questions from English and English Language Art subjects are not given in the dataset, so we randomly choose one fully marked student answer as the model answer to test ASSESS, SACS, and our ProtSi Network. \\
We first generated the word embeddings of few-shot datasets and saved them as TFRecordDataset files. The generated dataset for each question occupies about 100G of storage, and it takes about 40 minutes to generate using multi-processing.\\
We manually tune the hyperparameters and find that the learning rate as $10^{-3}$ and EPOCHS as 5 having the highest accuracy score. The monotonically decreasing function $t$ is designed as:
\begin{equation}
    t = \frac{EPOCHS - epoch}{EPOCHS + n}
\end{equation}
where $epoch$ is the current epoch index and n is 1 to avoid $t=1$. For each question, we use 80\% of the answers to generate training tasks and 20\% for validation tasks. 
\section{Hardware and Software Configuration}
All experiments are performed on Ubuntu Linux machine with 60-core AMD EPYC 7543 32-Core Processor with 360 GB RAM and 4 Nvidia A40 GPUs with 48 GM Memory. We use Python 3.8 and Tensorflow 2.5.0 with CUDA 11.2 as the deep learning framework.
\section{More Experimental Results}
\begin{figure}[H]
    \centering
    \includegraphics[width=3.5in]{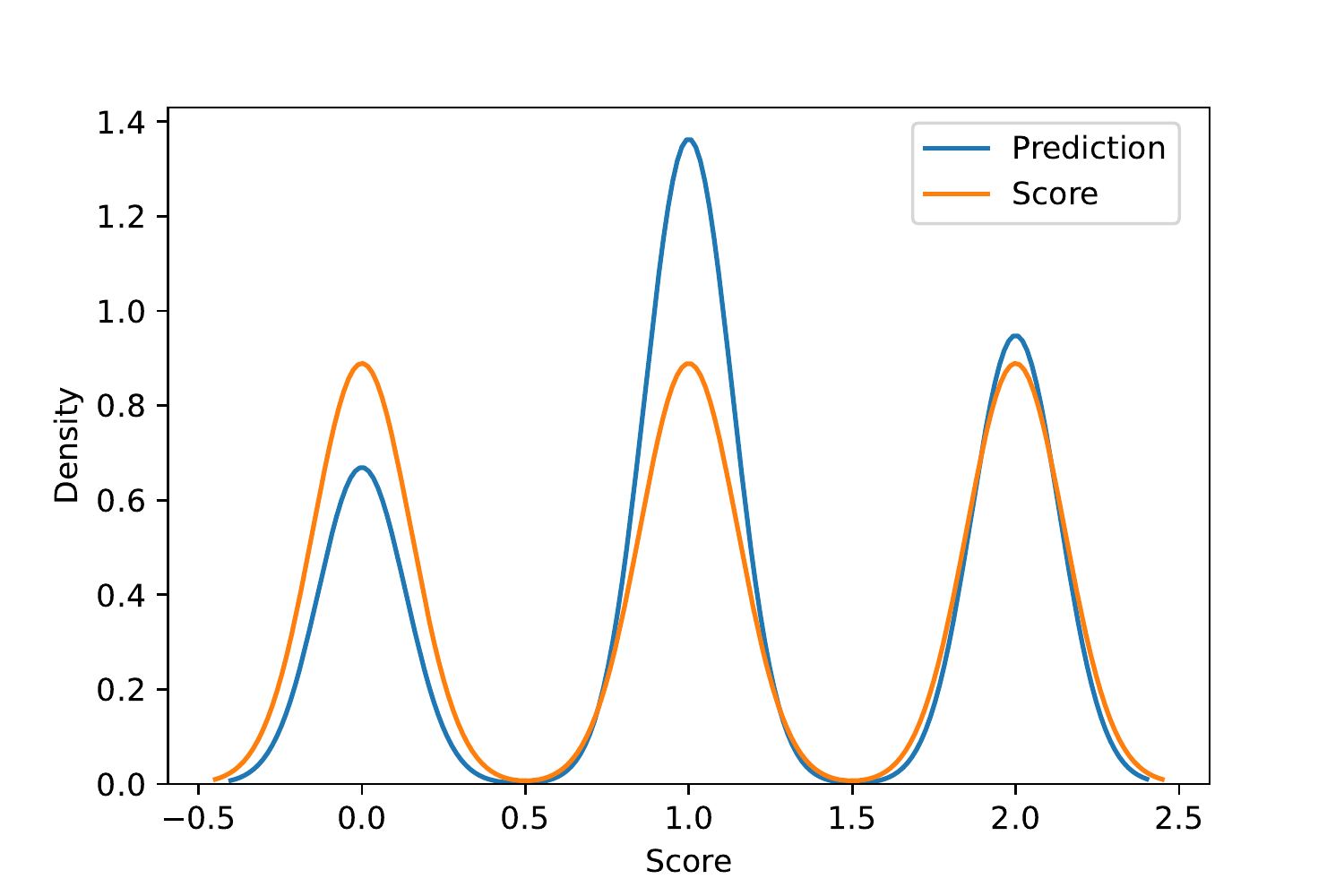}
    \caption{Predicted Score and real score distributions of $Q10$}
    \label{Q10}
\end{figure}

Figure \ref{Q10} presents the distributions of the predicted score and the real score. We observe that:
\begin{itemize}
    \item The distribution of predicted grades roughly matches the true distribution, especially for those full-mark answers.
    \item Some 0-point answers are misclassified as 1-point, illustrating that our model is not strict on scoring and the scoring strategy needs to be further defined and adjusted.
\end{itemize}

\end{document}